\documentclass[letterpaper, 10 pt, conference]{ieeeconf}  

\IEEEoverridecommandlockouts                              

\usepackage{amsmath,amsfonts}
\usepackage{algorithmic}
\usepackage{algorithm}
\usepackage{array}
\usepackage{textcomp}
\usepackage{stfloats}
\usepackage{url}
\usepackage{verbatim}
\usepackage{graphicx}
\usepackage{cite}
\usepackage{times}
\usepackage{epsfig}
\usepackage{amsmath}
\usepackage{amssymb}
\usepackage{cite}
\usepackage{booktabs}
\usepackage{multirow}
\usepackage{dsfont}
\usepackage{makecell}
\usepackage{comment}
\usepackage{color}
\usepackage{colortbl} 
\usepackage{hyperref}
\usepackage[table]{xcolor} 
\definecolor{light-gray}{gray}{0.9}

\title{\LARGE \bf
SGOR: Outlier Removal by Leveraging Semantic and Geometric Information for Robust Point Cloud Registration
}

\author{
Guiyu~Zhao, Zhentao Guo, and Hongbin~Ma,~\IEEEmembership{Senior~Member,~IEEE}
\thanks{This work was partially funded by the National
Key Research and Development Plan of China (No.
2018AAA0101000) and the National Natural Science
Foundation of China under grant 62076028 (Corresponding author: Hongbin~Ma)}
\thanks{Guiyu~Zhao, Zhentao Guo, and Hongbin~Ma are with the National Key Lab of Autonomous Intelligent Unmanned Systems, School of Automation, Beijing Institute of Technology, 100081, Beijing, P. R. China
(e-mail: 3120220906@bit.edu.cn, zt\_guo1230@163.com, mathmhb@bit.edu.cn).}
}

\begin{document}

\maketitle
\thispagestyle{empty}
\pagestyle{empty}

\begin{abstract}
In this paper, we introduce a new outlier removal method that fully leverages geometric and semantic information,  
to achieve robust registration.  
Current semantic-based registration methods only
use semantics for point-to-point or instance semantic
correspondence generation, which has two problems. 
First, these methods are highly dependent on the correctness of semantics.
They perform poorly in scenarios with incorrect semantics and sparse semantics. 
Second, 
the use of semantics is limited only to the correspondence generation,
resulting in bad performance in the weak geometry scene.
To solve these problems, on the one hand, we propose secondary ground segmentation 
and loose semantic consistency based on regional voting. 
It improves the robustness to semantic correctness by reducing the dependence on single-point semantics.
On the other hand, we propose semantic-geometric consistency for outlier removal, 
which makes full use of semantic information and significantly improves the quality of correspondences. 
In addition, a two-stage hypothesis verification is proposed, 
which solves the problem of incorrect transformation selection in the weak geometry scene.
In the outdoor dataset, our method demonstrates superior performance, boosting a 22.5 percentage points improvement in 
registration recall and achieving better robustness under various conditions.
\href{https://github.com/GuiyuZhao/SGOR}{Our code is available}.

\end{abstract}

\section{Introduction}

As an important task in 3D vision, point cloud registration has a wide range of 
applications, such as simultaneous localization and 
mapping~\cite{Segregator,10160810,10160697,10161425}, 
robot grasping~\cite{GMCR,10161193}, etc. 
There are many methods for point cloud registration, 
among which the most mainstream is the correspondence-based 
point cloud registration~\cite{IMF,yu2021cofinet,10160863,huang2021predator}. 
The correspondence-based method obtains a set of correspondences 
through feature extraction and matching, and then the 
transformation is estimated through SVD or RANSAC~\cite{fischler1981random}. 
However, the effectiveness of this method is heavily contingent on the quality of correspondences, 
particularly in scenarios characterized by low overlap, large scenes, and sparse features. 
Therefore, removing outliers to filter correspondences 
becomes a critical step in achieving robust and accurate registration.

\begin{figure}[!t]
        \centering{\includegraphics[scale=0.38]{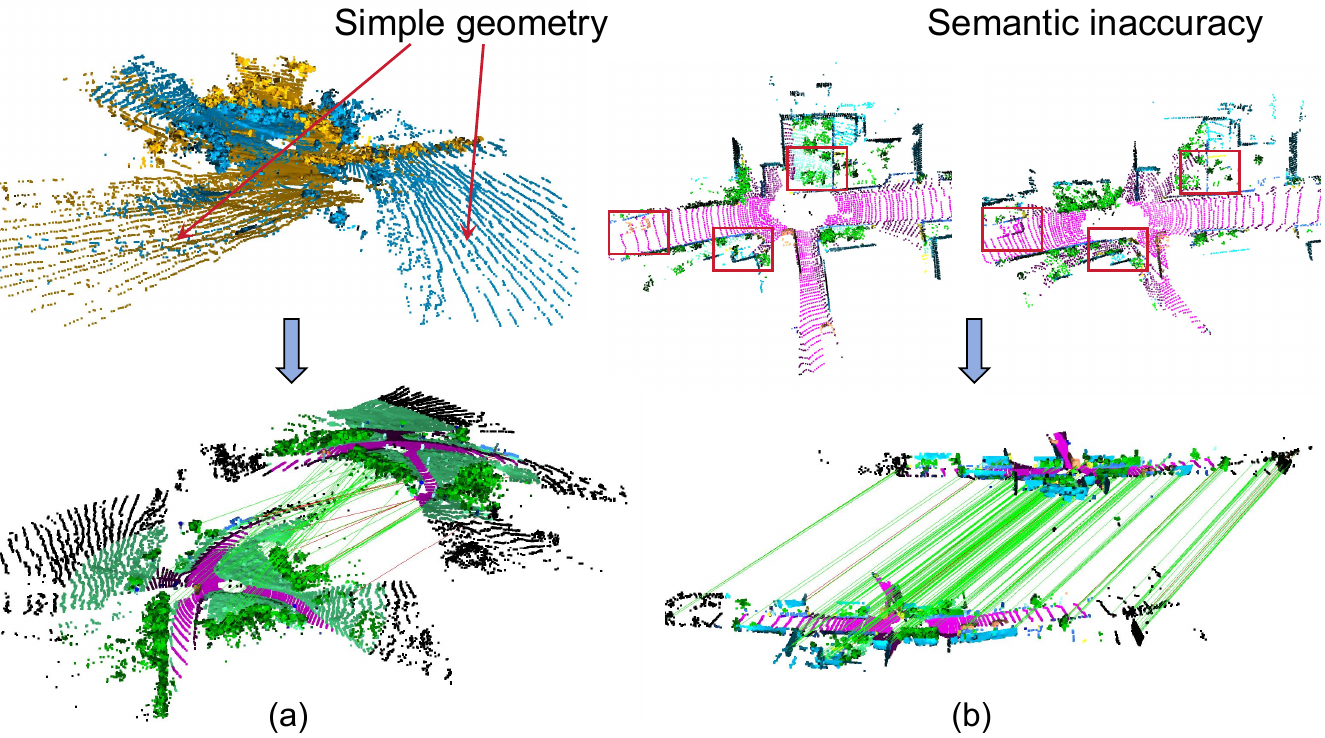}}  
        \caption{
                Our method can also achieve robust registration 
                even in the case of simple geometry and semantic inaccuracy. 
                The green line is the correct correspondence and 
                the red line is the wrong correspondence.
           }    
        \label{intro}
        \vspace{-5pt}
\end{figure}

In recent years, research in outlier removal for point cloud registration has gained significant momentum. 
This research can be categorized into two main approaches: 
learning-based methods and traditional geometric methods.
PointDSC~\cite{PointDSC} enhances spatial consistency 
within features through neural networks, resulting in 
correspondences of higher quality. 
SC2-PCR~\cite{chen2022sc2} introduces second-order spatial geometric consistency, 
notably enhancing the effectiveness of outlier removal. 
MAC~\cite{mac} incorporates the maximal clique of graph theory into point cloud registration, 
leading to more accurate registration. However, all these methods primarily 
focus on outlier removal by better capturing geometric consistency. 
These geometric-only methods perform poorly in flat, simple scenarios with lots of roads, streets, etc.

With the progress of semantic segmentation
research~\cite{cortinhal2020salsanext}, 
it has a great prospect to promote the improvement 
of other computer vision tasks. 
It is a good idea to use semantic information~\cite{Segregator,qiao2023pyramid} to 
improve geometry-only methods,
which solves the problem of insufficient geometric information.
However, they all have some defects, leading to failure in robust registration.
\textbf{First}, Segregator~\cite{Segregator} conducts instance-level semantic clustering
and Pagor~\cite{qiao2023pyramid} uses point-to-point semantic judgment.
Both of them are highly dependent on the accuracy of semantic segmentation, and
semantics error of the point will seriously affect the final result.
\textbf{Second}, 
Their use~\cite{Segregator, qiao2023pyramid} of semantics only is limited to correspondence 
generation, and the outlier removal and hypothesis verification are still only based on geometry,
which suffers from the same problems as geometric-only methods.
As shown in Fig.~\ref{intro},
our method solves these two problems.


In this paper, we propose a new method of using geometric and semantic information to 
achieve robust registration. 
\textbf{On the one hand}, we propose secondary ground segmentation and loose semantic 
consistency based on regional voting,
which reduces the dependence on semantic accuracy and is more robust to 
semantic errors.
\textbf{On the other hand}, we make full use of semantic 
prior information through semantic-geometric consistency and two-stage hypothesis verification 
based on the ground prior and solve the problem that the previous semantic-based 
methods~\cite{Segregator, qiao2023pyramid} still 
rely heavily on geometric information in outlier removal and hypothesis verification. 
Therefore, this enables our method to achieve robust registration 
in the face of geometrically deficient scenarios (large areas of ground).
Our method performs best in the outdoor dataset,
achieving more robust registration.

\begin{itemize}
        \setlength{\itemsep}{0pt}
        \setlength{\parsep}{0pt}
        \setlength{\parskip}{0pt}
        \item[$\bullet$] 
        We propose a new semantic-based idea to complete point cloud registration, 
        which achieves robust registration with a 22.5 pp improvement in registration recall.
        \item[$\bullet$] 
        By using secondary ground segmentation and
        loose regional semantic consistency, 
        our method is more robust to semantic accuracy and richness.
        \item[$\bullet$] 
        With the semantic-geometric consistency and verification based on ground prior,
        we fully leverage semantics in the outlier removal and hypothesis verification,
        solving the problem in the scenarios with weak geometry.

\end{itemize}

        

\section{Related Work}

\subsection{Outlier Removal for Point Cloud Registration}
Random Sampling Consensus (RANSAC)~\cite{fischler1981random, derpanis2010overview}, 
as a classical and 
effective correspondence estimation method, is also widely 
used today. However, it is difficult to converge when there 
are a large number of outliers. To address this challenge, 
a lot of works~\cite{zhou2016fast, choy2020deep, PointDSC, chen2022sc2, mac, jiang2023robust, zhao2024vrhcf} 
have been dedicated to outlier removal through geometric 
invariants. 
FGR~\cite{zhou2016fast} leverages the Geman-McClure cost to transform 
non-convex problems into convex ones, 
resulting in a significant enhancement of correspondence quality. 
DGR~\cite{choy2020deep} employs convolutional networks to 
predict the confidence levels of correspondences. 
PointDSC~\cite{PointDSC} introduces a spatial non-local consistency 
module that incorporates geometric consistency into features, 
subsequently filtering out outliers. 
GeoTransformer~\cite{qin2022geometric} utilizes a local-to-global 
registration module, providing an effective way for 
refining correspondences and achieving robust registration.
SC2-PCR~\cite{chen2022sc2} introduces second-order 
spatial geometric consistency, leading to substantial 
improvements in outlier removal performance.
Additionally, MAC~\cite{mac} introduces the maximal clique 
concept into point cloud registration, resulting in significant 
enhancements. However, 
These methods primarily focus on the utilization of geometric 
invariants for outlier removal while neglecting the 
significance of semantic priors, leading to not robust registration under
the condition of simple geometric structures and sparse features.

\subsection{Semantics for Enhancing Computer Vision Tasks}
With the development of deep learning and the proposal of 
foundation models~\cite{brown2020language, kirillov2023segment}, 
semantic segmentation tasks have developed rapidly. 
Using semantics as a prior to assist other computer vision tasks has 
become an efficient way. Previously, there has been a lot of work using 
semantic information for 3D reconstruction and SLAM. 
Menini et al.~\cite{menini2021real} use a deep learning-based 
method to introduce semantics 
into TSDF and improve the effect of indoor 3D reconstruction. 
Huang et al.~\cite{huang2021real} establish a semantic pose graph using semantic 
priors to achieve globally consistent 3D reconstruction. 
SuMa++~\cite{chen2019suma++} combines semantics into surfel-based mapping and then 
performs semantic ICP, which achieves the best performance in 
outdoor highway scenes. Segregator~\cite{Segregator} uses semantic and geometric 
information to cluster points into instance clusters, and match 
the instance clusters to achieve point cloud registration. 
Qiao et al.~\cite{qiao2023pyramid} use a pyramid semantic graph and cascaded 
gradient ascend method to achieve global registration. 
In the past two years, these few studies~\cite{Segregator, qiao2023pyramid} 
have introduced 
semantics into point cloud registration, and have achieved 
certain improvements, but there are many problems such as a narrow range of applicable scenarios 
and insufficient use of semantic information.

\section{Methodology}

\begin{figure*}[!t]
        \centering{\includegraphics[scale=0.65]{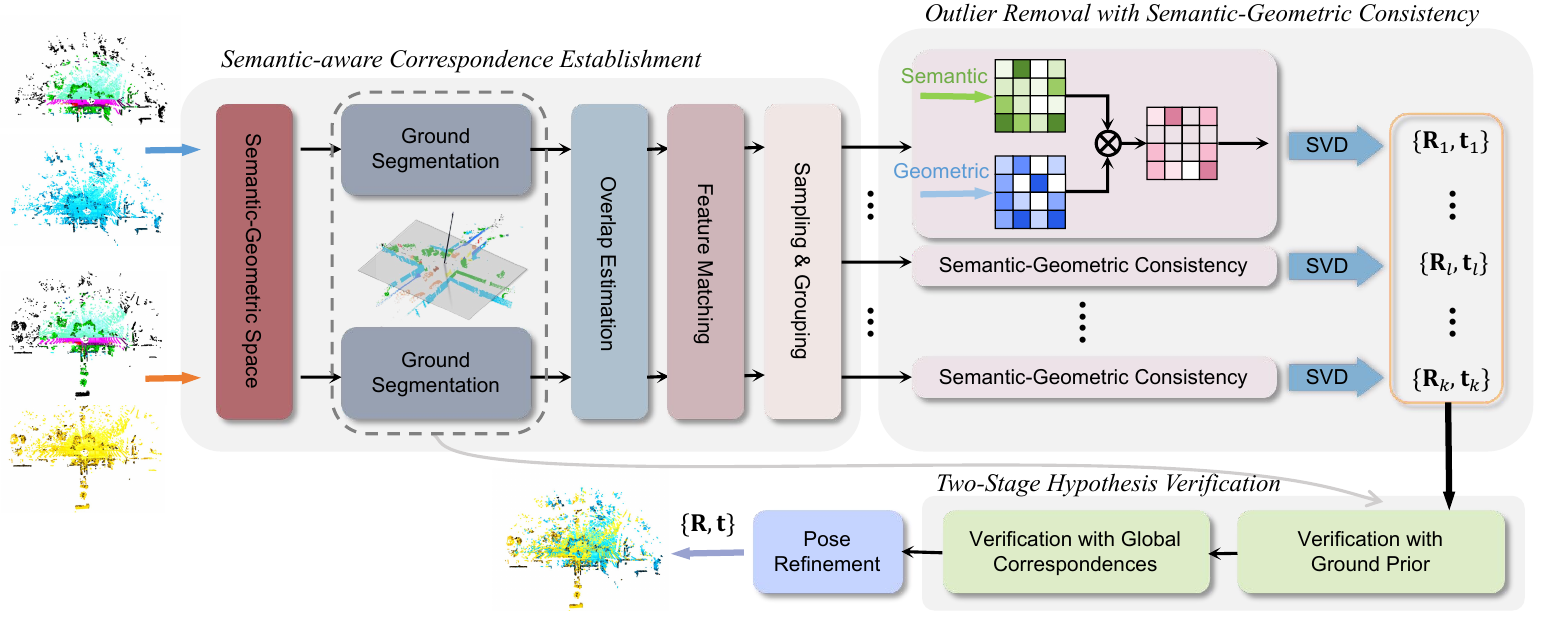}}  
        \caption{
        Pipeline of our proposed method. 
        First, we input the point cloud and its semantics to construct a semantic-geometric space $\mathbb{U}$. 
        Next, the secondary ground segmentation separates the point cloud into ground $\mathbf{U}_{\mathbf{P}g}$  
        and non-ground points $\mathbf{U}_{\mathbf{P}n}$ . 
        Subsequently, we estimate the overlap region $ \mathbf{S}_{\mathbf{P}_g}$ within the semantic space, 
        extract point features for feature matching, and establish the correspondences $\mathcal{G}$. 
        After that, we obtain local correspondences $\mathcal{G}^k$ by sampling 
        and grouping. Outliers within each local correspondence 
        are filtered based on semantic-geometric consistency, and  
        local transformations $\mathbf{T}_l$ are estimated. Finally, 
        the optimal transformation $\tilde{ \mathbf{R}}$ 
        is selected through the two-stage hypothesis verification.
           }
        \label{total}
        \end{figure*}

\subsection{Problem Formulation}
Given two point clouds $\mathbf{P}=\left\{\mathbf{p}_i \in \mathbb{R}^3 \mid i=1, \ldots, I\right\}$ and 
$\mathbf{Q}=\left\{\mathbf{q}_j \in \mathbb{R}^3 \mid j=1, \ldots, J\right\}$ with overlapping regions, 
point cloud registration is to align these two point clouds $\mathbf{P}$ and $\mathbf{Q}$ by a 
transformation $\mathbf T = \{\mathbf R,\mathbf t\}$ where $\mathbf{R} \in SO(3)$ and 
$\mathbf{t} \in \mathbb{R}^3$.
Furthermore, in correspondence-based methods, 
the correspondence set $\mathcal{I}$ is acquired through feature matching. 
Subsequently, these methods seek to determine the transformation $\mathbf{T}$ 
that minimizes the Euclidean distance between each point $\mathbf{p}_i$ 
in the transformed point cloud $\mathbf{T(P)}$ and its corresponding point $\mathbf{p}_j$ in the point cloud $\mathbf{Q}$

\begin{equation}
        \underset{\mathbf{R}\in SO(3), \mathbf{t}\in \mathbb{R}^3}{\arg \min } \sum_{\left({\mathbf{p}}_{i}, {\mathbf{q}}_{j}\right) \in \mathcal{I}} \left\|\mathbf{R} \cdot {\mathbf{p}}_{i}+\mathbf{t}-{\mathbf{q}}_{j}\right\|_2^2
\end{equation}
where this problem is usually solved by SVD. 
However, the effectiveness of this approach is heavily 
contingent on the accuracy of the correspondences. 
To enhance the accuracy of registration, 
we perform outlier removal to get a subset of 
more reliable correspondences $\mathcal{I}^{\prime}$, where $\mathcal{I}^{\prime} \subseteq \mathcal{I}$.
The overall framework of our approach is shown in Fig.~\ref{total}.

\subsection{Semantic-aware Correspondence Establishment}\label{Correspondence Establishment}
In the feature-based registration method, descriptors 
such as fast point feature histograms (FPFH)\cite{rusu2009fast} 
and fully convolutional geometric features (FCGF)\cite{choy2019fully} 
are extracted for feature matching. 
However, especially in scenarios with indistinct features, 
the inlier ratio of correspondences is quite low, 
leading to non-robust registration.
Furthermore, in the case of large outdoor scenes, 
a significant portion of the point cloud is occupied by 
points on the ground and road, which have indistinct 
features and limited discriminative capability. 
This leads to a significant challenge to establishing correspondences. 
To address these issues, we propose the semantic-aware correspondence establishment approach.

\textbf{Semantic-geometric space.} First, we acquire the semantic label $s_i$ of each point $\mathbf{p}_i$ through 
the mature semantic segmentation method~\cite{cortinhal2020salsanext} or directly using the semantic priors. 
By combining the 3D coordinates and semantic labels, 
a semantic-geometric space denoted as $\mathbb{U}= \mathbb{R}^3 \times \mathbb{S} $ is created,
where $\mathbb{S}$ represents the set of all possible semantic labels.
The semantic point cloud $\mathbf{U} \subseteq  \mathbb{U}$ 
serves as the input of our method where semantic 
point $(\mathbf{p}_i, s_i) \in \mathbf{U}$.

\begin{figure}[!t]
        \centering{\includegraphics[scale=0.28]{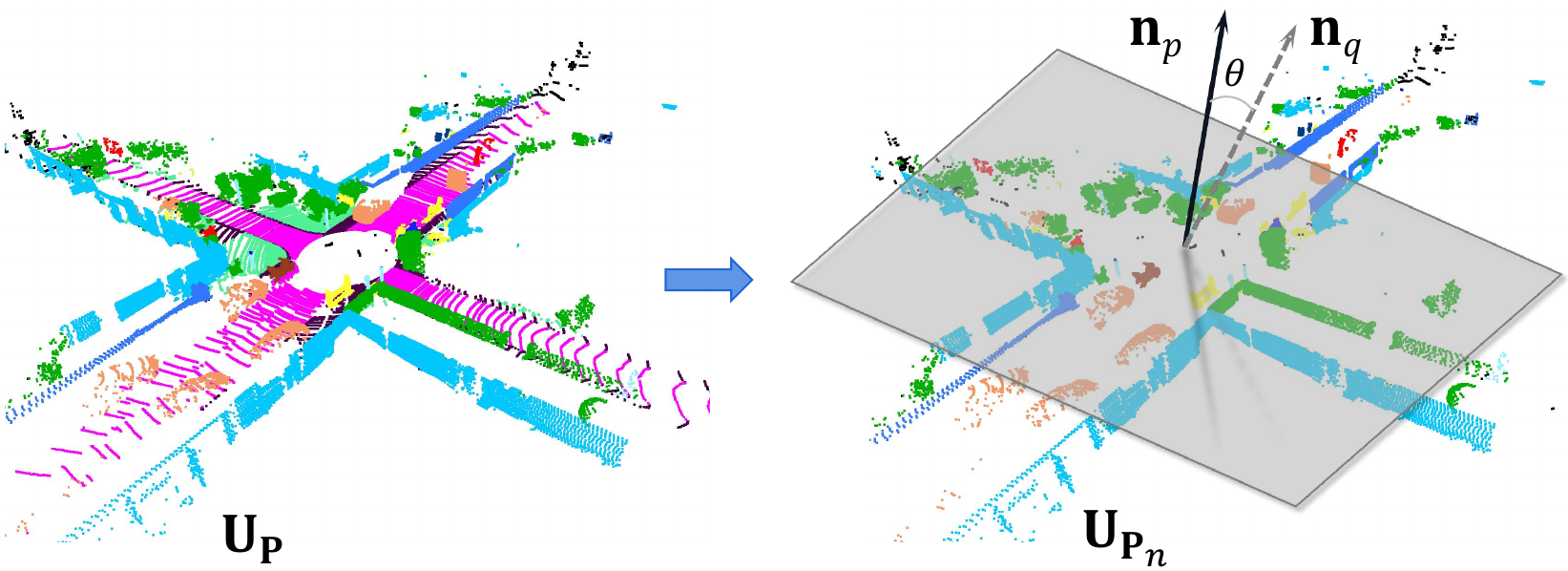}}  
        \caption{
                Ground segmentation and verification with ground prior.
           }    
        \label{ground}
        \vspace{-5pt}
\end{figure}

\textbf{Secondary  ground segmentation.} 
We represent the set of semantic labels pertaining to 
ground and road as $S_g$, 
while the set of labels for non-ground points is 
denoted as $S_n = \overline{S_g}$. 
By using semantic labels, we partition the semantic 
point cloud $\mathbf{U_P}$ into 
ground point clouds $\mathbf{U}_{\mathbf{P}g}$ and non-ground 
point clouds $\mathbf{U}_{\mathbf{P}_n}$ as follows:

\begin{equation}
        \mathbf{U}_{\mathbf{P}_g} = \{(\mathbf{p}_i, s_i)| s_i \in S_g, \forall (\mathbf{p}_i, s_i) \in \mathbf{U} \}.
\end{equation}
However, this segmentation method is highly dependent on single-point semantics.
To mitigate the influence of semantic errors of points, 
we conduct a secondary verification. 
We employ a simple SVD to extract planes $ax+by+cz+d=0$
from ground points $\mathbf{U}_{\mathbf{P}_g}$. 
Subsequently, we utilize the Euclidean distance from each point to the plane for secondary segmentation
and new ground points $\mathbf{U}_{\mathbf{P}_g}$ meet the following conditions:
\begin{equation}
        \frac{|a\mathbf{p}_i(x)+b\mathbf{p}_i(y)+c\mathbf{p}_i(z)+d|}{a^2+b^2+c^2} <\sigma_g
\end{equation}
where $\sigma_g$ is a distance threshold.
Then, we use SVD again on the new ground points $\mathbf{U}_{\mathbf{P}_g}$ to update the ground normal $\mathbf{n}_p \!\!\leftarrow \!\!(a^{\prime},b^{\prime},c^{\prime})$
This module helps reduce the reliance on single-point semantic information entirely.
It not only enhances processing efficiency by reducing the number of ground points 
but also filters out ground points with indistinctive features, 
thereby facilitating feature matching. 
Moreover, we also harness the ground normal $\mathbf{n}_p$ 
as a robust prior for guiding hypothesis verification in Section~\ref{Verification}.

In contrast to~\cite{huang2021predator} 
using the cross-attention mechanism to achieve overlap region estimation, 
we  obtain the intersection $\mathbf{S}_o$ 
of the semantic sets $ \mathbf{S}_{\mathbf{P}_n}$ and $ \mathbf{S}_{\mathbf{Q}_n}$ within the semantic space 
and then screen out the semantic overlap region 
\begin{equation}
        \begin{aligned}
        &\mathbf{U}_{\mathbf{P}_o} = \{(\mathbf{p}_i, s_i)| s_i \in \mathbf{S}_{\mathbf{P}_n} \cap \mathbf{S}_{\mathbf{Q}_n}, \forall (\mathbf{p}_i, s_i) \in \mathbf{U} \},\\
        &\mathbf{U}_{\mathbf{Q}_o} = \{(\mathbf{q}_i, s_i)| s_i \in \mathbf{S}_{\mathbf{P}_n} \cap \mathbf{S}_{\mathbf{Q}_n}, \forall (\mathbf{q}_i, s_i) \in \mathbf{U} \}.
\end{aligned}
\end{equation}

\textbf{Correspondences grouping.} 
We obtain the initial correspondences $\mathcal{G}$ by  
feature matching using FPFH (or FCGF) descriptor on the non-ground points. 
Different from~\cite{Segregator}, which directly clusters the correspondences 
into single-instance correspondences according to the semantic label, 
we use sampling and grouping methods to cluster the correspondence.
The clustered correspondences contain multiple semantic labels, 
which is more robust to semantic errors and more suitable for semantic consistency. 
First, correspondences $\mathcal{G}$ is sampled through  spectral
matching~\cite{leordeanu2005spectral} to get $L$ key correspondences $\mathcal{G}^k$, 
and then k-nearest neighbor (kNN) search is conducted on $\mathcal{G}^k$ based on source points 
to get $L$ sets of local correspondences $\mathcal{G}_l$ where $\mathcal{G}_l,l=1,\ldots, L$.


\subsection{Outlier Removal with Semantic-Geometric Consistency}
Despite the initial optimization of correspondences, a
considerable number of mismatches still persist.  
In this section, a semantic-geometric double consistency 
criterion is proposed to eliminate mismatches.

For local correspondences $\mathcal{G}_l$, 
geometric consistency matrix, and semantic 
consistency matrix are calculated respectively. 
For geometric consistency, we choose the transformation 
invariant of Euclidean space, the point pair 
distance, as the geometric consistency 
score $d_{ij}$ of one correspondence.
\begin{equation}
        d_{ij}=\big|\| \mathbf{p}_i- \mathbf{p}_j ||_2-||\mathbf{q}_i- \mathbf{q}_j\|_2\big|
\end{equation}
This score $d_{ij}$ is zeroed by giving a distance threshold $\sigma_d$. 
We get a local geometric consistency matrix $\mathbf{M}^g=[m^g_{ij}]_{k \times k}$ 
\begin{equation}
        m^g_{ij} = \mathds {1}(\frac{d^2_{ij}}{\sigma^2_d} -1  \leqslant 0)
\label{mg}
\end{equation}
where $i$ and $j$ are the indices of correspondences $\mathcal{G}_l$, and
$k$ is the number of correspondences in $\mathcal{G}_l$.
$\mathds {1}(\cdot)$ is the indicator function.
To overcome the vulnerability of local regions to outliers, 
we use the global geometric consistency information to guide the selection of local correspondences.
Therefore, 
we calculate the local-to-global geometric consistency matrix 
$\mathbf{M^{\prime}}=[m^{\prime}_{ij}]_{k \times w}$ 
where $i$ is the index of correspondences $\mathcal{G}_l$, $j$ 
is the index of correspondences $\mathcal{G}$,
and $w$ is the size of the correspondences $\mathcal{G}$.
Finally, we obtain the global-aware local geometric 
consistency matrix $\mathbf{M}^{*}_g$
\begin{equation}
        \mathbf{M}^{*}_g=\mathbf{M}_g\circ \left(\mathbf{M}_g^{\prime}{\mathbf{W}}_m{\mathbf{M}_g^{\prime}}^\top \right)
\end{equation}
where operator $\circ$ represents the element-wise product and
${\mathbf{W}}_m = [w^{\prime}_{ij}]_{w \times w}$ is the weight matrix of 
the consistency between correspondences, calculated as
\begin{equation}
        w^{\prime}_{ij}=\exp(-\frac{d_{ij}^2}{2\sigma_d^2}).
\end{equation}
With this distance weight ${\mathbf{W}}_m$, our geometric consistency 
matrix is more robust to anomalous correspondences.

\textbf{Analysis: Geometric meaning of Matrix $\mathbf{M}_g^{*}$.}
\emph{
The element in matrix $\mathbf{M}_g^{\prime}$ represents the consistency between a 
correspondence in $\mathcal{G}_l$ and a correspondence in $\mathcal{G}$. 
In cases where all weights $\mathbf{W}_m$ are equal to 1, 
$\mathbf{M}_g^{\prime}$ undergoes direct multiplication with 
${\mathbf{M}_g^{\prime}}^\top$, resulting in elements that denote 
the count of correspondences in $\mathcal{G}$ exhibiting consistency 
with two correspondences from $\mathcal{G}_l$. Subsequently, through 
element-wise multiplication with $\mathbf{M}_g$, the score of 
correspondences in $\mathcal{G}_l$ that fail to meet the consistency 
criteria is set to 0. In this case, 
the matrix quantifies the score of consistency achieved by the two sets 
of correspondences $\mathcal{G}_l$ within the global correspondences $\mathcal{G}$.
By introducing distance weights, correspondences that are distant from correspondences 
$\mathcal{G}_l$ are assigned lower weights, thereby enhancing their resistance to the influence of noise.}


Relying solely on geometric consistency may result in poor 
performance when confronted with a singular geometric 
structure of the actual scene, such as open roads or 
orderly blocks, leading to incorrectly filtering correspondences. 
To address this issue and achieve more robust registration, 
we introduce semantic consistency.
By comparing whether the 
semantics of the correspondences are the same,
we obtain a tight semantic 
consistency matrix of correspondences $\mathcal{G}_l$
\begin{equation}
        \mathbf{M}_s \!=\!\left[\begin{array}{ccc}
        (s^p_1,s^p_1)\odot (s^q_1,s^q_1)  & \cdots & (s^p_1,s^p_k)\odot (s^q_1,s^q_k) \\
        \vdots  & \ddots & \vdots \\
        (s^p_k,s^p_1)\odot (s^q_k,s^q_1)  & \cdots & (s^p_k,s^p_k)\odot (s^q_k,s^q_k) 
        \end{array}\right]_{k \times k}
\end{equation}
where the operator $\odot$ means to determine whether the 
two-dimensional vectors are the same.
Considering the accuracy of semantic segmentation, 
the semantic labels of two sets of correct 
correspondences may not necessarily match. 
Therefore, to reduce the reliance on semantic accuracy,
we loosen the constraint of the semantic consistency 
by neighbor-based semantic consistency.

We search the neighbors
$\mathcal{N}(\mathbf{p}_i)$ 
and $\mathcal{N}(\mathbf{q}_i)$
on the
corresponding points $\mathbf{p}_i$ and $\mathbf{q}_i$ within a radius.
Then, the semantic labels 
that have the highest proportion within the neighbor points are defined as
$\mathcal{S}(\mathbf{p}_i)$ 
and $\mathcal{S}(\mathbf{q}_i)$.
We calculate the neighbor-based semantic consistency matrix  $\mathbf{M}_s^{\prime}=[m^{s^{\prime}}_{ij}]_{k \times k}$
\begin{equation}
        m^{s^{\prime}}_{ij} = \left(\mathcal{S}(\mathbf{q}_i),\mathcal{S}(\mathbf{q}_j) \right)
        \odot
        \left(\mathcal{S}(\mathbf{p}_i),\mathcal{S}(\mathbf{p}_j) \right).
\end{equation}
We obtain the finall semantic 
consistency matrix
\begin{equation}
        \mathbf{M}_s^{*}=\mathbf{M}_s \otimes  \mathbf{M}_s^{\prime}
\end{equation}
where operator $\otimes$ represents the element-wise logical OR.

Finally, we embed semantic consistency into geometric consistency to get the semantic-geometric consistency
\begin{equation}
        \mathbf{M}^{*}=\mathbf{M}_s^{*} \circ \mathbf{M}_g^{*}.
\end{equation}
According to matrix $\mathbf{M}^{*}$, we apply the Top-k 
algorithm to filter out $k_1 (k_1 < k)$ pairs of the correspondences 
for each set of local correspondences $\mathcal{G}_l$.
\begin{equation}
    \mathcal{G}_l^{\prime}=\left\{ \left((\mathbf{p}_n, s^p_n),(\mathbf{q}_n, s^q_n) \right) \Big| 
        n=\underset{j\in [1,k]}{\text{topk} }( m_{lj}^{*}) \right\}
\end{equation}
After outlier removal, we get $L$ groups of the filtered correspondences, 
and then we use SVD to get $L$ candidate 
transformations.
\begin{equation}
        \mathbf{R}_l, \mathbf{t}_l=\min_{\mathbf{R}, \mathbf{t}} \sum\nolimits_{\left((\mathbf{p}_j, s^p_j),(\mathbf{q}_j, s^q_j) \right) \in \mathcal{G}_l^{\prime}}
         w_j^l\left\|\mathbf{R} {\mathbf{p}}_j+\mathbf{t}-{\mathbf{q}}_j\right\|_2^2
\end{equation}
where $ w_j^l$ is the weight of inlier pair $(\mathbf{p}_j, \mathbf{q}_j)$ in correspondences $\mathcal{G}_l^{\prime}$
\begin{equation}
        w_j^l=   m_{lj}^{*} / \sum_{j=1}^{k} m_{lj}^{*}
\end{equation}

\subsection{Two-Stage Hypothesis Verification}\label{Verification}

Traditional hypothesis verification based on inlier count often 
gets trapped in local optimality when encountering 
scenes with simple geometry and a large number of similar regions, 
leading to registration failure. To address this issue, 
we leverage the ground points $\mathbf{U}_{\mathbf{P}_g}$ segmented previously 
to perform ground normal verification, preemptively 
eliminating numerous local optimal solutions. Subsequently, 
we employ the truncation distance to select the optimal 
transformation $\tilde{ \mathbf{T}}$.
Therefore, we propose the two-stage verification for selecting transformation.

It can be seen from experience that in outdoor environments, 
within a certain distance, the normal direction of the ground plane remains consistent.
In Section~\ref{Correspondence Establishment}, 
we obtained the ground point cloud by ground segmentation. 
As shown in Fig.~\ref{ground}, by using SVD, we obtain the normal vector $\mathbf{n}_p$ of the ground
which serves as a robust prior to guide the selection of candidate transformations
\begin{equation}
        \mathbf{U}_p\mathbf{\Sigma }_p\mathbf{V}^\top_p=\sum_{\mathbf{p}_{i} \in \mathbf{P}_g} \left(\mathbf{p}_i - \boldsymbol{\mu} \right)\left(\mathbf{p}_i - \boldsymbol{\mu}\right)^\top
\end{equation}
where $\mathbf{P}_g$ is the ground part of the source point cloud
and $\boldsymbol{\mu} = \frac{1}{|\mathbf{P}_g|}\sum_{\mathbf{p}_{i} \in \mathbf{P}_g} \mathbf{p}_{i}$.
Ground normal $\mathbf{n}_p$ can be calculated as $\mathbf{V}_p[:,-1]$.
Similarly, the ground normal of the target point cloud is $\mathbf{n}_q = \mathbf{V}_q[:,-1]$.
The wrong candidate transformations are eliminated by judging 
whether the candidate transformations can align two ground normal vectors,
\begin{equation}
        \bigl |\cos\bigl( \mathbf{R}_l\mathbf{n}_p,\mathbf{n}_q \bigr)\bigl | < \sigma_\theta
\end{equation}
where $\sigma_\theta$ is an angle threshold. 
After ensuring that there are 
no obviously wrong transformations, 
the best candidate transformation is selected by using the average 
truncation distance among the filtered global correspondences
\begin{equation}
        \tilde{ \mathbf{R}}, \tilde{ \mathbf{t}}=\min _{\mathbf{R}_l, \mathbf{t}_l} \sum_{\left((\mathbf{p}_i, s^p_i),(\mathbf{q}_i, s^q_i) \right) \in \mathcal{G}^\prime} 
        TD\bigl(\mathbf{R}_l {\mathbf{p}}_i+\mathbf{t}_l,{\mathbf{q}}_i \bigr)
\end{equation}
where $\mathcal{G}^\prime= \mathcal{G}^\prime_1 \cup \mathcal{G}^\prime_2 \cup \ldots \cup \mathcal{G}^\prime_L$
and $TD(\cdot)$ is a truncated distance function, defined as
\begin{equation}
        TD({\mathbf{p}}_i, {\mathbf{q}}_i)=
\begin{cases}
        0.5 \sigma_d, & \|\mathbf{p}_i\!-\!\mathbf{q}_i \|_2\leqslant 0.5 \sigma_d\\
        \|\mathbf{p}_i-\mathbf{q}_i \|_2, & 0.5 \sigma_d < \|\mathbf{p}_i\!-\!\mathbf{q}_i \|_2 <2.5 \sigma_d\\
        2.5 \sigma_d, &  2.5 \sigma_d \leqslant \|\mathbf{p}_i\!-\!\mathbf{q}_i \|_2
\end{cases} 
\end{equation}

\subsection{Transformation Refinement}
To further improve the accuracy of registration, 
the global correspondences $\mathcal{G}$ are used to refine the transformation. 
We obtain the correspondences $\tilde{\mathcal{G}}$ that satisfies the best candidate transformation 
$\tilde{ \mathbf{T}}=\{\tilde{ \mathbf{R}}, \tilde{ \mathbf{t}}\}$ on correspondences $\mathcal{G}$
\begin{equation}
        \tilde{\mathcal{G}} = \left\{ \left((\mathbf{p}_i, s^p_i),(\mathbf{q}_i, s^q_i) \right) \in {\mathcal{G}} \bigwedge \left\|\tilde{ \mathbf{R}}{\mathbf{p}}_i\!+\!\tilde{ \mathbf{t}}\!-\!{\mathbf{q}}_i\right\|_2^2<\tau_1 \right\}
\end{equation}
On correspondences $\tilde{\mathcal{G}}$, we use SVD again to achieve a more accurate transformation 
${ \mathbf{T}}\{{ \mathbf{R}}, { \mathbf{t}}\}$ of the global registration.
\section{Experiments}
\begin{figure*}[!t]
        \centering{\includegraphics[scale=0.5]{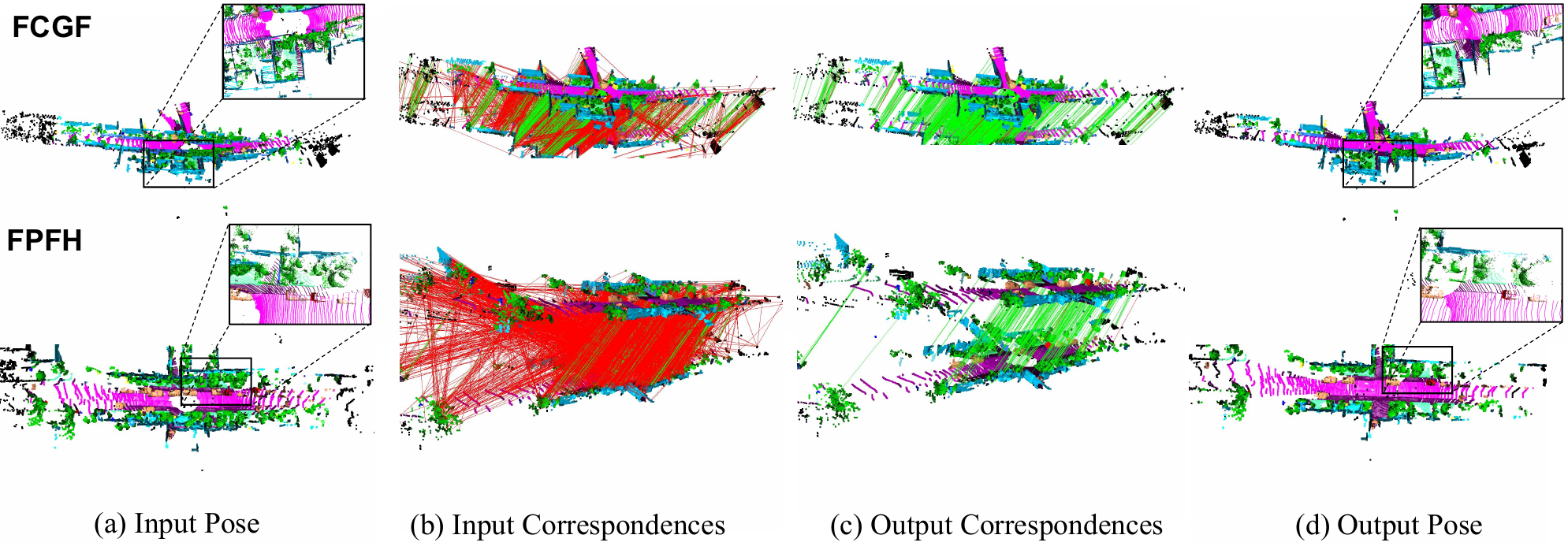}}  
        \caption{
                Correspondences result and registration result on KITTI dataset.
           }
        \label{qr1}
\end{figure*}

\subsection{Experimental Settings}

\subsubsection{Benchmark Dataset}
We conduct experiments with both indoor and outdoor datasets.
We have chosen the comprehensive KITTI dataset for outdoor scenes. 
Following~\cite{PointDSC,chen2022sc2,mac}, we conduct 
registration experiments on three sequences: 07, 08, and 09. 
In addition, we conduct indoor experiments on the 3DMatch dataset.
Following~\cite{chen2022sc2,mac}, we use 8 test scenes for evaluation.
\subsubsection{Evaluation Metric}

Following~\cite{chen2022sc2,mac}, we evaluate the registration results by registration recall (RR), 
isotropic rotation error (RE), and L2 translation error (TE).
Following~\cite{chen2022sc2}, we consider registration as accurate when RE $< 5^{\circ} $ 
and TE $<$ 60 cm for KITTI, and RE $< 15^{\circ} $ 
and TE $<$ 30 cm for 3DMatch. For outlier removal, we adopt three evaluation 
metrics~\cite{chen2022sc2}: inlier precision (IP), inlier recall (IR), and F1-score (F1). 
\subsubsection{Implementation Details}
The methods 
selected for experiments encompass the foremost outlier removal 
methods\cite{PointDSC,leordeanu2005spectral,fischler1981random,yang2020teaser,chen2022sc2,mac, choy2020deep, lee2021deep} 
as well as the most recent semantic-based registration 
methods~\cite{Segregator,qiao2023pyramid}. 
We implement methods~\cite{Segregator,qiao2023g3reg} under official guidelines.
For other methods,
following~\cite{chen2022sc2,mac,jiang2023robust}, 
we extract FPFH~\cite{rusu2009fast} and FCGF~\cite{choy2019fully} 
descriptors and sample 8000 points with features as inputs.
For KITTI, we directly predict semantics on point clouds by SalsaNext~\cite{cortinhal2020salsanext}.
However, 
Utilizing RGBD images for semantic prediction is more 
suitable for indoor scenes. Hence, we employ RGBD data 
for semantic prediction by ESANet~\cite{seichter2021efficient}, 
which is then projected into a semantic point cloud. 
To ensure fairness, we set the input of other methods to the point cloud formed by a single 
image projection, rather than the point cloud reconstructed by TSDF~\cite{zeng20173dmatch}.

\subsection{Evaluation on KITTI odometry}

\begin{figure}[!t]
        \centering{\includegraphics[scale=0.35]{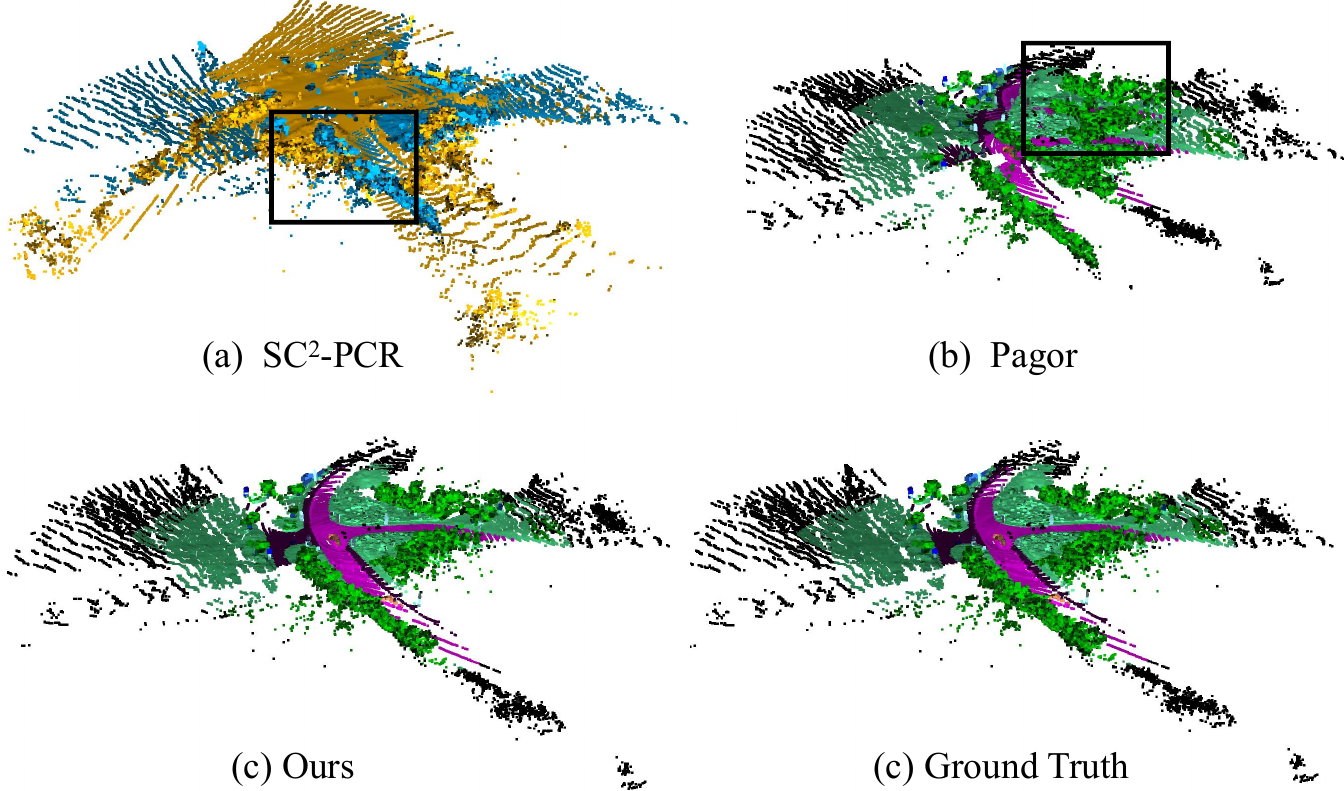}}  
        \caption{
        Qualitative results under indistinct geometric features
        and few semantic categories.
        Despite the state-of-the-art method~\cite{chen2022sc2} and the semantic 
        registration method~\cite{qiao2023pyramid} fail, 
        our approach still achieves robust registration.}
        \label{qr2}
        \vspace{-4pt}
\end{figure}

\begin{table}[htbp]
        \caption{Registration Result on KITTI Dataset.}
        \centering
        \resizebox{1.0\linewidth}{!}{
        \begin{tabular}{l|ccc|ccc|c}
        \toprule
        & \multicolumn{3}{c|}{\textbf{FPFH}} & \multicolumn{3}{c|}{\textbf{FCGF}} & \textbf{Time}\\
        Method & RR($\uparrow$)  & RE($\downarrow$) & TE($\downarrow$) & RR($\uparrow$)  & RE($\downarrow$) & TE($\downarrow$) &\textbf{Sec.} \\
        \midrule
        DHVR~\cite{lee2021deep}& -   &  -  & -  &98.20 &\underline{0.47} &20.54 &3.91\\
        DGR~\cite{choy2020deep}&78.21 &1.78 &32.12 &97.84 &0.48   &21.43   &1.57  \\
        PointDSC~\cite{PointDSC}&97.12 &0.59 &8.98 &97.48 &\underline{0.47}   &20.76   &0.30  \\
        VBReg~\cite{jiang2023robust}&96.04 &0.70 &15.29 &98.20 &\textbf{0.46}   &20.32   &0.33  \\
        \midrule
        SM~\cite{leordeanu2005spectral} &77.66 &0.63 &13.22 &96.22 &0.63   &21.27    &\textbf{0.08} \\
        RANSAC~\cite{fischler1981random} &- &- &- &75.68 &0.66   &27.89    &0.31 \\
        TEASER~\cite{yang2020teaser}& 90.27 &1.32 &15.74 &94.10 &0.55   &\underline{19.90}   &0.13  \\
        SC$^2$-PCR~\cite{chen2022sc2} &97.84 &\underline{0.58} &9.44 &\underline{98.38} &0.53   &20.29  &0.13   \\
        MAC~\cite{mac} &\underline{98.20} &0.60 &\underline{8.71} &\underline{98.38} &0.48   &20.19  &1.32   \\
        \rowcolor{light-gray} 
        Segregator~\cite{Segregator} &70.12 &1.52 &18.81 &- &-   &-    &0.14 \\
        \rowcolor{light-gray} 
        Pagor~\cite{qiao2023pyramid} &76.39&0.91&17.25& -& -  &-     &\textbf{0.08}\\
        \rowcolor{light-gray} 
        SGOR (\emph{ours}) &\textbf{98.92} &\textbf{0.51} &\textbf{7.50} &\textbf{98.74} &\textbf{0.46} &\textbf{19.77} &\underline{0.11}    \\
        \bottomrule
        \end{tabular}
        }
        \label{table:Registration kitti}
        \vspace{-5pt}       
\end{table}

\textbf{Registration result.}
As demonstrated  in Table~\ref{table:Registration kitti},
we conduct
comparative experiments with baselines
\cite{PointDSC, lee2021deep,leordeanu2005spectral,fischler1981random,yang2020teaser,chen2022sc2,mac, choy2020deep} 
on the KITTI dataset. 
The algorithms above the dividing line are learning-based methods, 
while the latter ones are non-learning methods. 
Furthermore, the highlighted rows represent methods 
leveraging semantic information. 
``-'' indicates the absence of results conducted under this specific condition in the official code.
From Table~\ref{table:Registration kitti}, it is evident that
whether using FPFH or FCGF descriptors, 
our method achieves the most robust and accurate registration. 
Figure~\ref{qr1} (d) presents the visualization of the registration results obtained using our method.
Compared with
geometric-only methods~\cite{PointDSC,chen2022sc2,mac,yang2020teaser}, 
our method introduces the fusion of semantic 
and geometric information, enabling more robust registration even in 
scenarios with indistinct geometric features,
as shown in Fig.~\ref{qr2}. Additionally, due to the 
verification with ground priors and semantic consistency, 
our registration exhibits lower TE and RE. 
Compared to the latest semantic registration 
methods~\cite{Segregator,qiao2023pyramid}, our 
method shows significant improvement, with RR increasing 
from 76.39\% to 98.92\%, and RE and TE decreasing from 0.71 
and 17.25 to 0.51 and 7.50, respectively.
Moreover, the running time of our method is 
relatively short, only 0.11s for 8000 correspondences.

\begin{table}[htbp]
        \caption{Correspondences Result on KITTI Dataset.}
        \centering
        \resizebox{1.0\linewidth}{!}{
        \begin{tabular}{l|ccc|ccc}
        \toprule
        & \multicolumn{3}{c|}{\textbf{FPFH}} & \multicolumn{3}{c}{\textbf{FCGF}} \\
        Method & IP($\uparrow$)  & IR($\uparrow$) & F1($\uparrow$) & IP($\uparrow$)  & IR($\uparrow$) & F1($\uparrow$)  \\
        \midrule
        DGR~\cite{choy2020deep}& 78.39 &54.12 &62.15 & 72.19 &78.06   & 75.13     \\
        PointDSC~\cite{PointDSC}&85.35 &81.08 &82.82 &81.25 &89.94   &85.01     \\
        VBReg~\cite{jiang2023robust}&\underline{91.68} &92.19 &\underline{91.88} &\textbf{95.31} & \textbf{96.84}  & \textbf{96.05}    \\
        \midrule
        SM~\cite{leordeanu2005spectral} &40.05 &\underline{93.98} &53.58 &\underline{93.34} & 15.13  &23.52     \\
        RANSAC~\cite{fischler1981random} &2.09 &15.72 &3.58 &57.90 &80.12 &65.89     \\
        TEASER~\cite{yang2020teaser}&82.56 &68.08 &73.98 &73.02 &67.99   &69.05     \\
        SC$^2$-PCR~\cite{chen2022sc2} &90.07 &{92.75} &91.27 &82.50 &91.52   &86.37     \\
        \rowcolor{light-gray} 
        SGOR (\emph{ours}) &\textbf{92.70} &\textbf{94.59} &\textbf{93.53} &{83.10} &\underline{91.56} &\underline{86.79}    \\
        \bottomrule
        \end{tabular}
        }
        \label{table:Correspondences kitti}

\end{table}

\textbf{Correspondences result.}
In addition to robust and accurate registration, 
our method offers another advantage: 
the ability to filter out correspondences. 
When using the FPFH descriptors, 
our method outperforms in the outlier removal.
As shown in Table~\ref{table:Correspondences kitti}, it achieves 92.70\% in IP and 94.59\% in IR, 
indicating that while effectively filtering out 
the majority of outliers, 
we manage to retain almost all correct correspondences. 
However, for the learning-based FCGF descriptors, 
the learning-based method VBreg~\cite{jiang2023robust} achieves the best performance. 
Still, our method shows significant improvement compared to non-learning methods~\cite{chen2022sc2}. 
As qualitative results are shown in Fig.~\ref{qr1},
our method accurately selects
correct correspondences even in cases 
with a high percentage of outliers, ultimately achieving precise registration.

\subsection{Evaluation on 3DMatch}

\begin{table}[htbp]
        \caption{Qualitative Result on 3DMatch.}
        \centering
        \resizebox{1.0\linewidth}{!}{
        \begin{tabular}{l|cccccc}
        \toprule
        & \multicolumn{6}{c}{\textbf{FPFH}} \\
        Method & RR($\uparrow$)  & RE($\downarrow$) & TE($\downarrow$) & IP($\uparrow$)  & IR($\downarrow$) & F1($\downarrow$) \\
        \midrule
        PointDSC~\cite{PointDSC}    &45.78 &2.82 &\textbf{8.53} &38.46 &40.17 &38.87\\
        VBReg~\cite{jiang2023robust}  &55.70 &3.17 &9.42 &46.33 &49.93 &47.69   \\
        \midrule
        SM~\cite{leordeanu2005spectral}   &31.25 &4.10 &11.09 &28.11 &51.84 &32.24   \\
        RANSAC~\cite{fischler1981random}  &6.41 &6.41 &15.24 &5.30 &34.06 &8.54  \\
        SC$^2$-PCR~\cite{chen2022sc2} &\underline{60.81} & \underline{2.94} &\underline{8.83} &\underline{51.26} &\textbf{57.27}   &\underline{53.74}    \\
        \rowcolor{light-gray} 
        SGOR (\emph{ours}) &\textbf{65.01} &\textbf{2.51} &{9.02} &\textbf{55.12} &\underline{55.90} &\textbf{55.51}   \\
        \bottomrule
        \end{tabular}
        }
        \label{table:Registration 3DMatch}
        \vspace{-5pt}       
\end{table}

\textbf{Qualitative results.}
To evaluate the generalization ability of our approach, we conduct experiments on indoor scenes, 
and the results are presented in Table~\ref{table:Registration 3DMatch}. It is worth noting that these experiments are performed 
using the point clouds made by a single depth map, which typically exhibit incomplete geometric structures 
and low overlap, leading to overall performance inferior to that reported in~\cite{chen2022sc2,mac}. 
Table~\ref{table:Registration 3DMatch} demonstrates 
that our SGOR method outperforms SC2-PCR~\cite{chen2022sc2} in terms of RR by using FPFH descriptors, with improvements 
of 4.2 pp. This enhancement is attributed to our method's ability to effectively 
address challenges such as incomplete geometry and low overlap, resulting in more robust registration. 
Furthermore, our approach also performs well in outlier removal on the 3DMatch dataset. 
Compared to SC2-PCR~\cite{chen2022sc2}, our method achieves higher IP, indicating superior outlier filtering capabilities.



\subsection{Robustness Test}

\begin{table}[htbp]
        \caption{Robustness Test under Different Error Thresholds.}
        \centering
        \resizebox{1.0\linewidth}{!}{
        \begin{tabular}{l|ccc|ccc}
        \toprule
        & \multicolumn{3}{c|}{\textbf{FPFH}} & \multicolumn{3}{c}{\textbf{FCGF}} \\
        Method & easy  & medium & hard & easy  & medium & hard  \\
        \midrule
        PointDSC~\cite{PointDSC}&97.12&95.14 &\underline{67.03} &97.48 &74.59   &\underline{23.06}     \\
        VBReg~\cite{jiang2023robust}&96.04 &88.65 &27.75 &98.20 &75.68   &22.52     \\
        \midrule
        RANSAC~\cite{fischler1981random} &- &- &- &75.68 &43.78   &6.49     \\
        SM~\cite{leordeanu2005spectral} &77.66 &69.73 &39.28 &96.22 &\underline{76.04}   &\textbf{23.60}     \\
        SC$^2$-PCR &\underline{97.84} &\underline{96.94} &59.82 &\underline{98.38} &75.32   &\underline{23.06}     \\
        \rowcolor{light-gray} 
        Segregator~\cite{Segregator} &70.12 &54.10 &10.68 &- &-  & -     \\
        \rowcolor{light-gray} 
        Pagor~\cite{qiao2023pyramid} &76.39 &58.01 &9.18 &- &-  & -    \\
        \rowcolor{light-gray} 
        SGOR (\emph{ours}) &\textbf{98.92} &\textbf{97.66} &\textbf{76.40} &\textbf{98.74} &\textbf{77.66}  &\textbf{23.60}     \\
        \bottomrule
        \end{tabular}
        }
        \label{Robustness Test}
       
\end{table}

\textbf{Robustness to different thresholds.}
Different tasks have different error requirements.
We conduct a robustness test 
by applying three distinct error 
thresholds: easy (5$^{\circ}$, 60cm), 
medium (5$^{\circ}$, 30cm), and hard (2$^{\circ}$, 10cm).
The results, as shown in Table~\ref{Robustness Test}, 
indicate that our method achieves the highest
success rate under all three different conditions. 
Specifically, under the strict criteria of RE $< 2^{\circ} $ 
and TE $< 10$ cm, our method still achieves a 76.40\% success rate.

\begin{figure}[htbp]
        \centering{\includegraphics[scale=0.088]{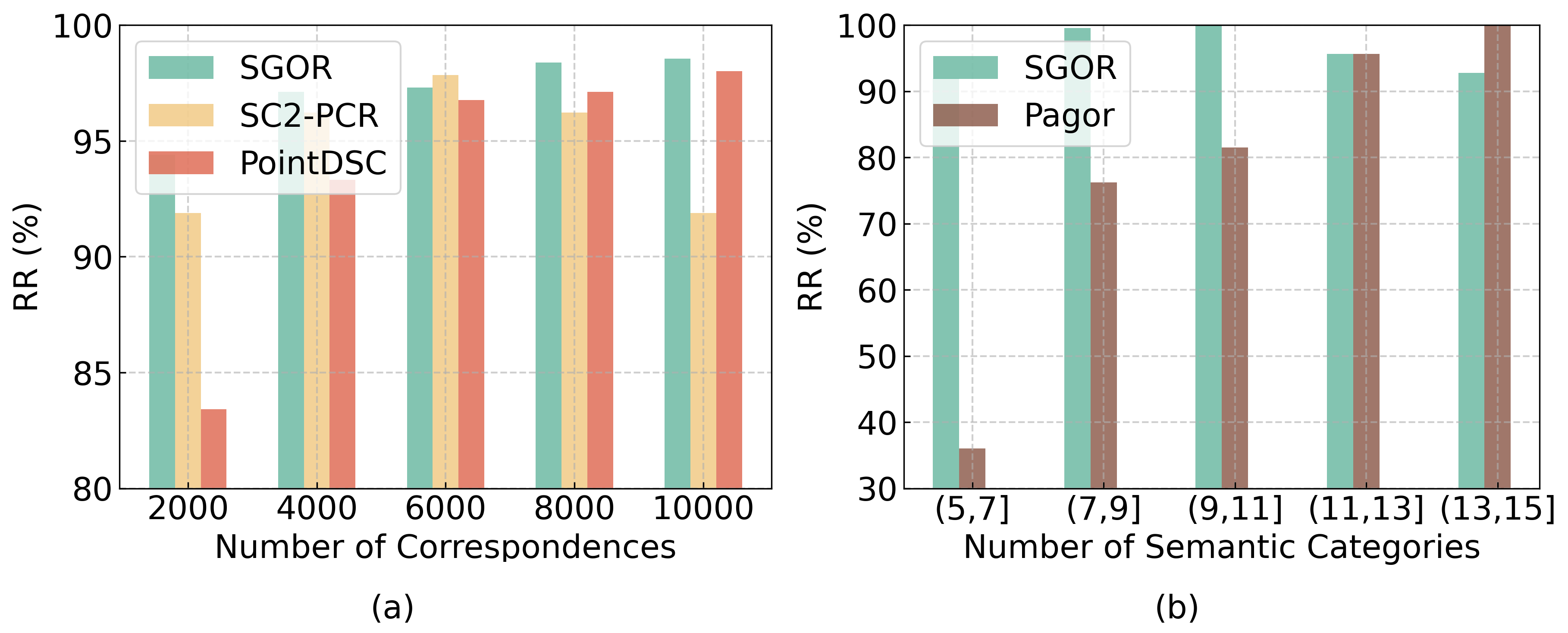}}  
        \caption{
        (a): Registration Recall under different numbers of correspondences;
        (b): Registration Recall under different semantic conditions.
        }
        \label{number of correspondences}
        \vspace{-10pt}
\end{figure}

\textbf{Robustness to correspondence quantity.}
We conduct random sampling of correspondences to get 
different numbers of correspondences. As shown in Fig.~\ref{number of correspondences} (a), 
SC2-PCR~\cite{chen2022sc2} shows a significant decrease in RR 
with too many or too few correspondences. PointDSC~\cite{PointDSC} fails to achieve 
robust registration when there are fewer correspondences. In contrast, 
our method exhibits good robustness to the number of correspondences 
and maintains high performance.

\textbf{Robustness to semantic label quantity.}
As shown in Fig.~\ref{number of correspondences} (b), according to the number of effective 
semantic categories, 
we divide the KITTI dataset into four types. 
when the 
number of semantic categories decreases,
the most advanced semantic-based method~\cite{qiao2023pyramid} 
experiences a significant decrease in RR, dropping from 90\% to 36\%. 
In contrast, our method consistently maintains a high level of performance even when poor semantics
like Fig.~\ref{qr2}.

\textbf{Robustness to semantic label quality.}
We further validate the robustness of our method to semantic quality through experiments. 
To simulate poor semantic prediction, we randomly substitute semantic labels for 50\% points in the point cloud. 
Under this condition, our method is compared with Pagor~\cite{qiao2023pyramid} and Segregator~\cite{Segregator}. 
After noise experiments, our approach achieves 
a 91.21\% RR in the high-noise scenarios. Compared to the original decrease of only 7.61 pp, 
our method exhibits a much lower decline than Pagor's decrease of 42.90 pp 
and Segregator's decrease of 15.09 pp.

\subsection{Ablation Study}
In this section, we conduct ablation experiments on the 
KITTI dataset to validate the effectiveness of individual modules. 
The experimental results are presented in Table~\ref{Ablation}
where * indicates the default settings of our SGOR.

\textbf{Preprocessing of correspondences.}
We perform preprocessing of correspondences, involving the 
filtering of ground points and estimation of potential overlapping 
regions.  A comparison between methods No.1 and 2 clearly demonstrates that, 
with correspondence preprocessing, our approach 
achieves a significant improvement in RR and a reduction in  registration errors.
Furthermore, due to filtering many useless points, the processing speed is
improved.

\textbf{Secondary ground segmentation.}
The plane segmentation module is essential to our approach. 
Its accuracy will directly affect the feature matching 
and candidate transformation screening.
We use the simple plane estimation only by predicted semantics to replace the 
secondary ground segmentation.
As can be seen from rows No.3, and 4 in Table \ref{Ablation}, 
the performance of the ablation model has been significantly diminished,
which illustrates the importance of our plane segmentation method.

\textbf{Using Semantic information.}
From rows No.5, and 7 in Table \ref{Ablation}, 
it is evident that incorporating semantic-geometric information 
into the geometric-only method yields noticeable improvements,
increasing RR from 91.35\%/97.66\% to 98.92\%/98.74\%. 
The results underscore the significance and effectiveness of our 
semantic-geometric method.
To further highlight the superiority of our loose  
regional semantic consistency, we replace it with a strict point-to-point semantic 
consistency like~\cite{qiao2023pyramid}. It can be seen from Table~\ref{Ablation} (6) and (7)
that our method reduces the dependence on semantic accuracy and 
achieves more accurate and robust registration.


\textbf{Two-stage verification.}
We have enhanced the hypothesis verification
for ground verification, utilizing  the point cloud of the ground
with indistinct features. As demonstrated in 
experiments No. 8 and 9, it is evident that this ground 
verification makes our method more robust, improving  RR by 13.01\%.



\begin{table}[htbp]
        \scriptsize
        \setlength{\tabcolsep}{1.5pt}
        \centering
        \caption{Ablation Study on KITTI Dataset. }
        \resizebox{1.0\linewidth}{!}{
        \begin{tabular}{l|l|ccc|ccc}
        \toprule
        & &\multicolumn{3}{c|}{\textbf{FPFH}} & \multicolumn{3}{c}{\textbf{FCGF}} \\
        No.  &Methods        & RR($\uparrow$)  & RE($\downarrow$) & TE($\downarrow$) & RR($\uparrow$)  & RE($\downarrow$) & TE($\downarrow$) \\
        \midrule
     
        1)         &W/o preprocessing of correspondences       &89.01    &0.55    &9.32  &97.48 &0.54 &20.29  \\
        2)         &W/ preprocessing of correspondences*       &\textbf{98.92} &\textbf{0.51} &\textbf{7.50} &\textbf{98.74} &\textbf{0.46} &\textbf{19.77}     \\
        \midrule  
        3)         &Ground segmentation only by semantics        &90.21&0.67 &9.25 &97.14 &0.75 &20.00  \\
        4)         &Secondary ground segmentation*             &\textbf{98.92} &\textbf{0.51} &\textbf{7.50} &\textbf{98.74} &\textbf{0.46} &\textbf{19.77}     \\
        \midrule
        5)         &Geometric-only SGOR        &91.35 &0.53 &9.34 &97.66 &0.51 &20.26 \\
        6)         &Semantic-hard SGOR       &95.81 &0.63 &9.41 &95.78 &0.69 &21.00 \\
        7)         &Semantic-geometric SGOR*  &\textbf{98.92} &\textbf{0.51} &\textbf{7.50} &\textbf{98.74} &\textbf{0.46} &\textbf{19.77}     \\
        \midrule  
        8)         &Verification without ground prior       &85.91&0.57 &9.97 &96.90 &0.62 &20.26  \\
        9)         &Two-stage verification*       &\textbf{98.92} &\textbf{0.51} &\textbf{7.50} &\textbf{98.74} &\textbf{0.46} &\textbf{19.77}     \\
        

        \bottomrule
        \end{tabular}
        }
        \label{Ablation}
     
     \end{table}

       

\section{CONCLUSION}
This paper presents a novel method for outlier removal that leverages semantic information. 
This method effectively filters out erroneous correspondences, resulting in a 
significantly more robust and accurate registration. By harnessing semantic 
information and ground priors of outdoor scenes,
we address the challenge of incorrect correspondences and registration failures in 
difficult scenarios marked by indistinct features and few semantic
labels. 
In our future work, we intend to 
develop a lightweight network 
for extracting both semantics and features to cooperate with our SGOR.

\vspace{5pt}

\bibliographystyle{IEEEtran}
\bibliography{IEEEabrv,icra}

\end{document}